\documentclass{article}
\usepackage{graphicx} 

\usepackage[english]{babel}

\usepackage[a4paper]{geometry}

\usepackage{multirow}
\usepackage{amsmath}
\usepackage{array}
\usepackage{adjustbox}
\usepackage{colortbl}
\usepackage{booktabs}

\usepackage[colorlinks=true, allcolors=blue]{hyperref}

\title{Intent Classification for Bank Chatbots through LLM Fine-Tuning}
\author{Bibiána Lajčinová \textsuperscript{(1)}, Patrik Valábek \textsuperscript{(1), (3)}, Michal Spišiak \textsuperscript{(2)}}
\date{\footnotesize\textsuperscript{\textbf{(1)}}Slovak National Supercomputing Centre, Bratislava, Slovak Republic\\
 \textsuperscript{\textbf{(2)}}nettle, s.r.o., Bratislava, Slovak Republic\\
 \textsuperscript{\textbf{(3)}}Institute of Information Engineering, Automation, and Mathematics, Slovak University of Technology in Bratislava, Slovak Republic\\
}

\begin{document}

\maketitle

\begin{abstract}
This study evaluates the application of large language models (LLMs) for intent classification within a chatbot with predetermined responses designed for banking industry websites. Specifically, the research examines the effectiveness of fine-tuning SlovakBERT compared to employing multilingual generative models, such as \textit{Llama 8b instruct} and \textit{Gemma 7b instruct}, in both their pre-trained and fine-tuned versions. The findings indicate that SlovakBERT outperforms the other models in terms of in-scope accuracy and out-of-scope false positive rate, establishing it as the benchmark for this application.
\end{abstract}

\section{Introduction}
The advent of digital technologies has significantly influenced customer service methodologies, with a notable shift towards integrating chatbots for handling customer support inquiries. This trend is primarily observed on business websites, where chatbots serve to facilitate customer queries pertinent to the business’s domain. These virtual assistants are instrumental in providing essential information to customers, thereby reducing the workload traditionally managed by human customer support agents.

In the realm of chatbot development, recent years have witnessed a surge in the employment of generative artificial intelligence technologies to craft customized responses. Despite this technological advancement, certain enterprises continue to favor a more structured approach to chatbot interactions. In this perspective, the content of responses is predetermined rather than generated on-the-fly, ensuring accuracy of information and adherence to the business's branding style. The deployment of these chatbots typically involves defining specific classifications known as intents. Each intent correlates with a particular customer inquiry, guiding the chatbot to deliver an appropriate response. Consequently, a pivotal challenge within this system lies in accurately identifying the user’s intent based on their textual input to the chatbot.

\section{Problem Description and Our Approach}
\label{sec:problemdescr}
This work is a joint effort of Slovak National Competence Center for High-Performance Computing and nettle, s.r.o., which is a Slovakia-based start-up focusing on natural language processing, chatbots, and voicebots. HPC resources of Devana system were utilized to handle the extensive computations required for fine-tuning LLMs. The goal is to develop a chatbot designed for an online banking service. 
\vspace{0.5cm}

In frameworks as described in the introduction, a predetermined precise response is usually preferred over a generated one. Therefore, the initial development step is the identification of a domain-specific collection of intents crucial for the chatbot's operation and the formulation of corresponding responses for each intent. 
These chatbots are often highly sophisticated, encompassing a broad spectrum of a few hundreds of distinct intents.
For every intent, developers craft various exemplary phrases that they anticipate users would articulate when inquiring about that specific intent. These phrases are pivotal in defining each intent and serve as foundational training material for the intent classification algorithm.

Our baseline proprietary intent classification model, which does not leverage any deep learning framework, achieves a 67\% accuracy on a real-world test dataset described in the next section.
The aim of this work is to develop an intent classification model using deep learning, that will outperform this baseline model.

We present two different approaches for solving this task.
The first one explores the application of Bidirectional Encoder Representations from Transformers (BERT), evaluating its effectiveness as the backbone for intent classification and its capacity to power precise response generation in chatbots. The second approach employs generative large language models (LLMs) with prompt engineering to identify the appropriate intent with and without fine-tuning the selected model.  

\subsection{Data}
\label{sec:data}
Our training dataset consists of pairs (text, intent), wherein each text is an example query posed to the chatbot, that triggers the respective intent. This dataset is meticulously curated to cover the entire spectrum of predefined intents, ensuring a sufficient volume of textual examples for each category.

In our study, we have access to a comprehensive set of intents, each accompanied by corresponding user query examples. We consider two sets of training data: a ``simple'' set, providing 10 to 20 examples for each intent, and a ``generated'' set, which encompasses 20 to 500 examples per intent, introducing a greater volume of data albeit with increased repetition of phrases within individual intents. 

These compilations of data are primed for ingestion by supervised classification models. This process involves translating the set of intents into numerical labels and associating each text example with its corresponding label, followed by the actual model training.

Additionally, we utilize a test dataset comprising approximately 300 (text, intent) pairs extracted from an operational deployment of the chatbot, offering an authentic representation of real-world user interactions. All texts within this dataset are tagged with an intent by human annotators. This dataset is used for performance evaluation of our intent classification models by feeding them the text inputs and comparing the predicted intents with those annotated by humans. All of these datasets are proprietary to nettle, s.r.o., so they cannot be discussed in more detail here.

\subsection{Evaluation}
In this article, the models are primarily evaluated based on their in-scope accuracy using a real-world test dataset comprising 300 samples. Each of these samples belongs to the in-scope intents on which the models were trained. Accuracy is calculated as the ratio of correctly classified samples to the total number of samples. For models that also provide a probability output, such as BERT, a sample is considered correctly classified only if its confidence score exceeds a specified threshold. Throughout this article, accuracy refers to this in-scope accuracy.

As a secondary metric, the models are assessed on their out-of-scope false positive rate, where a lower rate is preferable. For this evaluation, we use artificially generated out-of-scope utterances. The model is expected either to produce a low confidence score below the threshold (for BERT) or generate an 'invalid' label (for LLM, as detailed in their respective sections).

\subsection{Approach 1: BERT-based Intent Classification}
\subsubsection{SlovakBERT}
\label{sec:slkBert}
Since the data at hand is in the Slovak language, the choice of a model with Slovak understanding was inevitable. Therefore, we have opted for a model named SlovakBERT \cite{slovakbert}, which is the first publicly available large-scale Slovak masked language model.

Multiple experiments were undertaken by fine-tuning this model before arriving at the top-performing model. These trials included adjustments to hyperparameters, various text preprocessing techniques, and, most importantly, the choice of training data. 

Given the presence of two training datasets with relevant intents (``simple'' and ``generated''), experiments with different ratios of samples from these datasets were conducted. The results showed that the optimal performance of the model is achieved when training on the ``generated'' dataset.

After the optimal dataset was chosen, further experiments were carried out, focusing on selecting the right preprocessing for the dataset. The following options were tested:
\begin{itemize}
    \item turning text to lowercase,
    \item removing diacritics from text, and
    \item removing punctuation from text.
\end{itemize}
Additionally, combinations of these three options were tested as well. Given that the leveraged SlovakBERT model is case-sensitive and diacritic-sensitive, all of these text transformations impact the overall performance.

Findings from the experiments revealed that the best results are obtained when the text is lowercased and both diacritics and punctuation are removed.

Another aspect investigated during the experimentation phase was the selection of layers for fine-tuning. Options to fine-tune only one quarter, one half, three quarters of the layers, and the whole model were analyzed (with variations including fine-tuning the whole model for the first few epochs and then a selected number of layers further until convergence). The outcome showed that the average improvement achieved by these adjustments to the model’s training process is statistically insignificant. Since there is a desire to keep the pipeline as simple as possible, these alterations did not take place in the final pipeline.

Every experiment trial underwent assessment three to five times to ensure statistical robustness in considering the results. 

The best model produced from these experiments had an average accuracy of 77.2\% with a standard deviation of $0.012$.

\subsubsection{Banking-Tailored BERT}
Given that our data contains particular banking industry nomenclature, we opted to utilize a BERT model fine-tuned specifically for the banking and finance sector. However, since this model exclusively understands the English language, the data had to be translated accordingly.

For the translation, DeepL API\footnote{\href{https://developers.deepl.com/docs}{www.deepl.com}} was employed. Firstly, training, validation, and test data was translated. Due to the nature of the English language and translation, no further correction (preprocessing) was done to the text, as discussed in~\ref{sec:slkBert}. Subsequently, the model's weights were fine-tuned to enhance performance.

The fine-tuned model demonstrated promising initial results, with accuracy slightly exceeding $70\%$. Unfortunately, further training and hyperparameter tuning did not yield better results. Other English models were tested as well, but all of them produced similar results. Using a customized English model proved insufficient to achieve superior results, primarily due to translation errors. The translation contained inaccuracies caused by the 'noisiness' of the data, especially within the test dataset.

\subsection{Approach 2: LLMs for Intent Classification}
As mentioned in Section~\ref{sec:problemdescr}, in addition to fine-tuning SlovakBERT model and other BERT-based models, the use of generative LLMs for the intent classification was explored too. Specifically, instruct models were selected for their proficiency in handling instruction prompts and question-answering tasks.
Since there are not open-source instruct model exclusively trained for the Slovak language, several multilingual models were selected: \textit{Gemma 7b instruct} \cite{gemma} and \textit{Llama3 8b instruct} \cite{llama3modelcard}. For comparison, we also include results for the closed-source OpenAI's \textit{gpt-3.5-turbo} model under the same conditions.

Similarly to \cite{parikh2023exploringzerofewshottechniques}, we use LLM prompts with intent names and descriptions to perform zero-shot prediction. The output is expected to be the correct intent label. Since the full set of intents with their descriptions would inflate the prompt too much, we use our baseline model to select only top 3 intents. Hence the prompt data for these models was created as follows:
Each prompt includes a sentence (user's question) in Slovak, four intent options with descriptions, and an instruction to select the most appropriate option. The first three intent options are the ones selected by the baseline model, which has a Top-3 recall of 87\%. The last option is always `invalid' and should be selected when neither of the first three matches the user's question or the input intent is out-of-scope. Consequently, the highest attainable in-scope accuracy in this setting is 87\%.

\subsubsection{Pre-trained LLM Implementation}
Initially, a pre-trained LLM implementation was utilized, meaning a given instruct model was leveraged without fine-tuning on our dataset. A prompt was passed to the model in the user's role, and the model generated an assistant's response. 

To improve the results, prompt engineering was employed too. It included subtle rephrasing of the instruction; instructing the model to answer only with the intent name, or with the number/letter of the correct option; or placing the instruction in the system's role while the sentence and options were in the user's role.

Despite these efforts, this approach did not yield better results than SlovakBERT's fine-tuning. However, it helped us identify the most effective prompt formats for fine-tuning of these instruct models. Also, these steps were crucial in understanding the models' behaviour and response pattern, which we leveraged in fine-tuning strategies of these models.

\subsubsection{LLM Optimization through Fine-Tuning}
The prompts that the pre-trained models reacted best to were used for fine-tuning of these models. Given that LLMs do not require extensive fine-tuning datasets, we utilized our ``simple'' dataset as detailed in section~\ref{sec:data}. The model was then fine-tuned to respond to the specified prompts with the appropriate label names.

Due to the size of the chosen models, parameter efficient training (PEFT) \cite{peft} strategy was employed to handle the memory and time issues. PEFT updates only a subset of parameters, while “freezing” the rest, therefore reducing the number of trainable parameters. Specifically, the Low-Rank Adaptation (LoRA) \cite{lora} approach was used.

To optimize performance, various hyperparameters were tuned too, including learning rate, batch size, \textit{lora alpha} parameter of the LoRA configuration, the number of gradient accumulation steps, and chat template formulation.

Optimizing language models involves high computational demands, necessitating the use of HPC resources to achieve the desired performance and efficiency. The Devana system, with each node containing 4 NVidia A100 GPUs with 40GB of memory each, offers significant computational power. In our case, both models we are fine-tuning fit within the memory of one GPU (full size, not quantized) with a maximum batch size of 2. Although leveraging all 4 GPUs in a node would reduce training time and allow for a larger overall batch size (while maintaining the same batch size per device), for benchmarking purposes and to guarantee consistency and comparability of the results, we conducted all experiments using 1 GPU only.

These efforts led to some improvements in models' performances. Particularly for \textit{Gemma 7b instruct} in reducing the number of false positives. On the other hand, while fine-tuning \textit{Llama3 8b instruct}, both metrics (accuracy and the number of false positives) were improved. However, neither \textit{Gemma 7b instruct} nor \textit{Llama3 8b instruct} models outperformed the capabilities of the fine-tuned SlovakBERT model.

With \textit{Gemma 7b instruct}, some sets of hyperparameters resulted in high accuracy but also a high false positive rate, while others led to lower accuracy and low false positive rate. Search for a set of hyperparameters bringing balanced accuracy and false positive rate was challenging. The best-performing configuration achieved an accuracy slightly over 70\% with a false positive rate of 4.6\%. Compared to the model’s performance without fine-tuning, fine-tuning only slightly increased the accuracy, but dramatically reduced the false positive rate by almost 70\%.

With \textit{Llama3 8b instruct}, the best-performing configuration achieved an accuracy of 75.1\% with a false positive rate of 7.0\%. Compared to the model’s performance without fine-tuning, fine-tuning significantly increased the accuracy and also halved the false positive rate.

\subsubsection{Comparison with a Closed-Source Model}
To benchmark our approach against a leading closed-source LLM, we conducted experiments using OpenAI's\footnote{\href{https://platform.openai.com/docs/overview}{www.openai.com}} \textit{gpt-3.5-turbo}. We employed identical prompt data for a fair comparison and tested both the pre-trained and fine-tuned versions of this model. Without fine-tuning, \textit{gpt-3.5-turbo} achieved an accuracy of 76\%, although it exhibited a considerable false positive rate. After fine-tuning, the accuracy improved to almost 80\%, and the false positive rate was considerably reduced.

\section{Results}
In our initial strategy, involving fine-tuning SlovakBERT model for our task, we achieved average accuracy of 77.2\% with a standard deviation of $0.012$, representing an increase of 10\% from the baseline model's accuracy.

Fine-tuning banking-tailored BERT on translated dataset showcased the final accuracy slightly under 70\%, which outperforms the baseline model, however it does not surpass the performance of fine-tuned SlovakBERT model. 

Subsequently, we experimented with pre-trained (but not fine-tuned with our data) generative LLMs for our task. While these models showed promising capabilities, their performance was inferior to that of the SlovakBERT fined-tuned for our specific task. Therefore, we proceeded to fine-tune these models, namely \textit{Gemma 7b instruct} and \textit{Llama3 8b instruct}. 
The fine-tuned \textit{Gemma 7b instruct} models demonstrated a final accuracy comparable to the banking-tailored BERT, while fine-tuned \textit{Llama3 8b instruct} performance was slightly worse than the SlovakBERT fined-tuned. Despite extensive efforts to find the configuration surpassing the capabilities of the SlovakBERT model, these attemps were unsuccessful, establishing the SlovakBERT model as our benchmark for performance.

All results are displayed in Table \ref{tab:results}, including the baseline proprietary model and a closed-source model for comparison.

\begin{table}[h!]
\centering
\begin{tabular}{lcc}
\toprule
\textbf{Model Name} & \textbf{In-scope Accuracy} & \textbf{Out-of-scope FPR} \\
\midrule
Baseline proprietary model & 67.6 & 22.5 \\
SlovakBERT fine-tuned & 77.2 & ~6.3 \\
Banking-tailored BERT & 68.5 & ~4.0 \\
Gemma 7b instruct pre-trained & 69.5 & 73.6 \\
Gemma 7b instruct fine-tuned & 70.6 & ~4.6 \\
Llama3 8b instruct pre-trained & 65.5 & 14.1 \\
Llama3 8b instruct fine-tuned & 75.1 & ~7.0 \\
gpt-3.5-turbo pre-trained & 76.6 & 32.4 \\
gpt-3.5-turbo fine-tuned & 79.5 & ~4.3 \\
\bottomrule
\end{tabular}
\caption{Percentage comparison of models' in-scope accuracy and out-of-scope false positive rate.}
\label{tab:results}
\end{table}

\section{Conclusion}
The goal of this study was to find an approach leveraging a pre-trained language model (fine-tuned or not) as a backbone for chatbot for banking industry. The data provided for the study consisted of pairs of text and intent, where the text represents user's (customer's) query and the intent represents the triggered intent. 

Several language models were experimented with, including SlovakBERT, banking-tailored BERT and generative models \textit{Gemma 7b instruct} and \textit{Llama3 8b instruct}. After experimentations with the dataset, fine-tuning configurations and prompt engineering; fine-tuning SlovakBERT emerged as the best approach yielding final accuracy slightly above 77\%, which represents a 10\% increase from the baseline's models accuracy, demonstrating its suitability for our task.

In conclusion, our study highlights the efficacy of fine-tuning pre-trained language models for developing a robust chatbot with accurate intent classification. Moving forward, leveraging these insights will be crucial for further enhancing performance and usability in real-world banking applications. 
\section{Acknowledgments}
The research results were obtained with the support of the Slovak National competence centre for HPC, the EuroCC 2 project and Slovak National Supercomputing Centre under grant agreement 101101903-EuroCC 2-DIGITAL-EUROHPC-JU-2022-NCC-01.

This work is a joint effort of Slovak National Competence Center for High-Performance Computing\footnote{\href{https://www.nscc.sk/}{www.nscc.sk}} and nettle, s.r.o.\footnote{\href{https://www.nettle.ai}{www.nettle.ai}}, a Slovakia-based start-up focusing on natural language processing, chatbots, and voicebots.

\bibliographystyle{plain}
\bibliography{references}

\end{document}